\documentclass[a4paper]{article}

\usepackage[english]{babel}
\usepackage[utf8]{inputenc}
\usepackage[T1]{fontenc}

\usepackage{listings}

\usepackage{lmodern}
\usepackage[autostyle]{csquotes}

\usepackage[a4paper,top=3cm,bottom=2cm,left=3cm,right=3cm,marginparwidth=1.75cm]{geometry}

\usepackage{amsmath}
\usepackage{graphicx}
\usepackage[colorinlistoftodos]{todonotes}
\usepackage{authblk}
\usepackage{rotating}
\usepackage{csvsimple}
\usepackage{float}

\usepackage[colorlinks=true, allcolors=blue]{hyperref}
\usepackage{rotating}

\title{\huge The Three Ghosts of Medical AI: Can the Black-Box Present Deliver?}
\author[1*]{Thomas P. Quinn}
\author[1]{Stephan Jacobs}
\author[1]{Manisha Senadeera}
\author[1]{Vuong Le}
\author[2]{Simon Coghlan}

\affil[1]{\footnotesize Applied Artificial Intelligence Institute, Deakin University, Geelong, Australia}
\affil[2]{\footnotesize Centre for AI and Digital Ethics, School of Computing and Information Systems, University of Melbourne, Melbourne, Australia

* \textit{contacttomquinn@gmail.com}
}
\date{}

\Affilfont{\fontsize{4}{4}}

\begin{document}
\maketitle

\begin{abstract}
Our title alludes to the three Christmas ghosts encountered by Ebenezer Scrooge in \textit{A Christmas Carol}, who guide Ebenezer through the past, present, and future of Christmas holiday events. Similarly, our article will take readers through a journey of the past, present, and future of medical AI. In doing so, we focus on the crux of modern machine learning: the reliance on powerful but intrinsically opaque models. When applied to the healthcare domain, these models fail to meet the needs for transparency that their clinician and patient end-users require. We review the implications of this failure, and argue that opaque models (1) lack quality assurance, (2) fail to elicit trust, and (3) restrict physician-patient dialogue. We then discuss how upholding transparency in all aspects of model design and model validation can help ensure the reliability of medical AI.
\end{abstract}

\maketitle

\section{Introduction}

We seem to hear almost everyday about how a new algorithm will revolutionize healthcare \cite{topol_high-performance_2019}. Artificial intelligence (AI), by which we roughly mean the application of computers to perform tasks that normally require human intelligence, has fast become a cornerstone of medical research. The zeitgeist suggests that the healthcare industry -- including the management, diagnosis, and treatment of patients -- is on the eve of a total transformation.
AI-based changes promise to transform clinical decision-making and clinician workflow, usher in direct-to-consumer medical services, and even provide robot-aided healthcare. But despite the promise of AI, there are present obstacles. One clinician noted:

\begin{quote}
After hearing for several decades that computers will soon be able to assist with difficult diagnoses, the practicing physician may well wonder why the revolution has not occurred. Skepticism at this point is understandable. Few, if any, programs currently have active roles as consultants to physicians.
\end{quote}
%https://www.nejm.org/doi/pdf/10.1056/NEJM198208193070808?articleTools=true
These words are timely and relevant -- except they were written over 30 years ago \cite{schwartz_artificial_1987}. While discussions about the AI revolution take place, hospitals continue to operate largely as ``business-as-usual''. Computers have indeed transformed healthcare, but mostly as a replacement for paper, not as a replacement for human intelligence and expertise. When a patient visits a medical clinic, they are almost invariably examined, diagnosed, and treated without any support from AI.

Will medical AI deliver on its promise? Maybe, but perhaps not without a change to our current research priorities. While domain knowledge guided the development of early medical AI, modern machine learning algorithms often require no such guidance whatsoever. The new models are powerful, but often intrinsically opaque: they fail to meet the needs for transparency that their clinician and patient end-users require. This intrinsic opacity \cite{wang_should_2019} has been attributed to a mismatch between algorithmic computation and the human demands for reasoning and interpretation \cite{burrell_how_2016}, leaving clinicians and patients in the dark about how the model works. (It is worth noting that an `extrinsic' lack of transparency can occur when clinicians or patients do not have access to an understanding of an algorithm due to, for example, proprietary secrets.)

\begin{figure}[H]
\centering
\scalebox{1}{
\includegraphics[width=(.75\textwidth)]{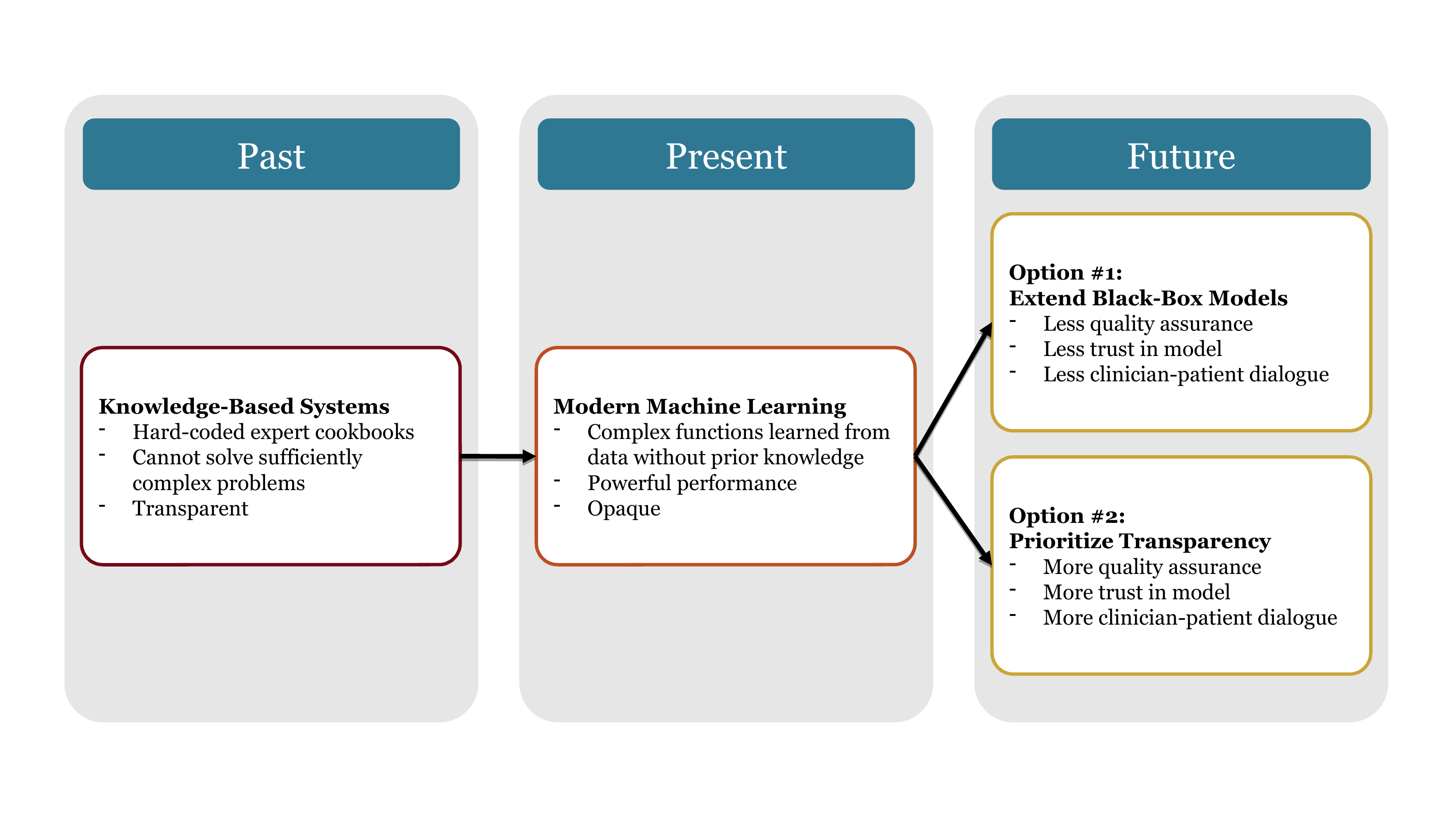}}
\caption{The field of artificial intelligence (AI) in healthcare began with knowledge-based systems that leveraged expert knowledge to hard-code predictive functions. Modern machine learning makes use of large data sets to learn the predictive functions automatically. Will future methods extend these powerful but opaque models? Or will they prioritize transparency?}
\label{fig:options}
\end{figure}

Like Ebenezer Scrooge's encounter with the three ghosts of Christmas in \textit{A Christmas Carol}, this article will give a tour of the past and present state of medical AI, and lay out the future options (see Figure~\ref{fig:options}). We are at a crossroads. Do we continue to extend these powerful but intrinsically opaque models? Or do we prioritize transparency and develop new algorithms that align with the needs of the clinicians and patients who will use them?
We review the implications of ``black-box'' AI, and argue that opaque models (1) lack quality assurance, (2) fail to elicit trust, and (3) restrict clinician-patient dialogue. We posit that healthcare should mandate transparency in all aspects of model design and model validation in an effort to help ensure the reliability of medical AI.%%, unless post-hoc explanations can be theoretically justified and empirically validated.

%This article begins with a brief survey of the history of medical AI, then presents a critical view of modern ``black-box'' machine learning. %We argue that opaque models cannot achieve the goals of medical AI because 
%We discuss how opaque models lack quality assurance, fail to elicit trust, and restrict clinician-patient dialogue, that make them inappropriate for the kinds of high-stakes decision-making required by medical applications.

%%%%%%%%%%%%%%%%%%%%%%%%%%%%%%%%%%%%%%%%%%%%%%%%%%%%
%%%%%%%%%%%%%%%%%%%%%%%%%%%%%%%%%%%%%%%%%%%%%%%%%%%%
\section{The Ghost of AI Past}
%%%%%%%%%%%%%%%%%%%%%%%%%%%%%%%%%%%%%%%%%%%%%%%%%%%%
%%%%%%%%%%%%%%%%%%%%%%%%%%%%%%%%%%%%%%%%%%%%%%%%%%

%
The field of AI originated in 1943 with a proposed model of an artificial neuron capable of reasoning and learning \cite{mcculloch_logical_1943}. This field developed alongside computers for several years with new programs that solved abstract ``toy problems'' through formal logic (e.g., proving geometric theorems or solving algebra story problems) \cite{russell_artificial_2010}. In the 1960s, the paradigm shifted away from logical inference to \textit{knowledge-based systems}, ushering in the first AI boom \cite{buchanan_very_2005}. Knowledge-based systems work by articulating expert knowledge through a series of ``hard-coded'' if-then statements, resulting in a kind of specialist ``cookbook'' \cite{russell_artificial_2010}. For the first time, AI began to solve ``real problems'', while the use of logic granted the model some transparency by providing a trace of the inference steps \cite{holzinger_what_2017}. This resulted in a lot of hype, especially within healthcare. In a letter to the JAMA editors in 1960, one professor wrote, ``Those of us who work closely with computers know them...as \emph{tremendously powerful tools of analysis, capable of making decisions as complex as man can describe}'' (emphasis in original) \cite{galler_value_1960}.
The MYCIN system, developed in the early 1970s, was the first major attempt to emulate clinical reasoning. However, knowledge-based systems soon proved ineffective for solving sufficiently complex problems \cite{schwartz_artificial_1987}. 
Excitement about AI ended abruptly when James Lighthill released a commissioned report that accused researchers of overstating the abilities of AI, resulting in a substantial drop in funding \cite{haenlein_brief_2019}. The AI boom thus ended in disillusionment, and so began what is now called the ``first AI winter''.

A second AI boom began in 1980s when a commercial expert system brought huge financial gains for the Digital Equipment Corporation, saving an estimated 40 million dollars per year by 1986 \cite{russell_artificial_2010}. Other companies invested in expert system technology with strong profit motives, and AI went from a few million dollar industry to a 2 billion dollar industry in less than 10 years \cite{russell_artificial_2010}.
Around this time, the INTERNIST-1 system, originally conceived in parallel to MYCIN, had reached maturity. This system attempted to construct and evaluate medical hypotheses through a diagnosis-ranking algorithm, but likewise proved ineffective for real-world use \cite{schwartz_artificial_1987,barnett_computer_1982}. Skepticism about AI in medicine took hold early, but soon became pervasive in other industries too.
By the late 1980s, the expert system market began to shrink as new research stopped transitioning into industry \cite{hendler_avoiding_2008}, and the cost of maintenance began to outweigh the utility \cite{partridge_scope_1987}. Again, the AI boom ended in disillusionment, and so began what is now called the ``second AI winter''.

Knowledge-based systems have since been largely abandoned for a new approach to AI called \textit{machine learning}, which seeks to relate variables to outcomes through complex non-linear functions learned directly from the data (without the necessity for prior expert knowledge).
Machine learning research continued to develop throughout the 1990s-2000s, as the popularity of deep artificial neural networks (ANN) took hold \cite{jensen_rise_2011}. Although there were some early applications of ANN to medical decision-making in the early 1990s, these early algorithms were not necessarily more accurate than conventional regression models \cite{tu_advantages_1996}. Today, we find ourselves in the midst of a third AI boom, propelled by technical innovations that make it possible to identify signals across space and time \cite{fukushima_neocognitron_1980,waibel_phoneme_1989,hochreiter_long_1997} (e.g., to recognize a face no matter where it occurs in an image), as well as the fast computers and ``big data'' needed to train \textit{deep} neural networks \cite{lecun_deep_2015}. As before, the momentum has carried over into medicine.

%%%%%%%%%%%%%%%%%%%%%%%%%%%%%%%%%%%%%%%%%%%%%%%%%%%%
%%%%%%%%%%%%%%%%%%%%%%%%%%%%%%%%%%%%%%%%%%%%%%%%%%%%
\section{The Ghost of AI Present}
%%%%%%%%%%%%%%%%%%%%%%%%%%%%%%%%%%%%%%%%%%%%%%%%%%%%
%%%%%%%%%%%%%%%%%%%%%%%%%%%%%%%%%%%%%%%%%%%%%%%%%%%%

Deep learning has rekindled excitement in AI by solving many difficult problems, especially in image and language processing. Yet, there are several ways in which AI could fail to deliver utility in healthcare. On one hand, extrinsic problems like legal liability and data scarcity could prevent the adoption of an ideal model. On the other hand, the lack of transparency in AI can be an intrinsic problem that, if unaddressed, threatens another AI winter in the healthcare domain.

Much of modern machine learning developed for low-stakes decisions, like online advertising, where model performance is more important than model intelligibility \cite{rudin_why_2019}. Although deep learning models are powerful, they tend to lack transparency: even if a model makes the correct prediction, it may be impossible to know and understand \textit{why}. This is due in part to the immense complexity of the `neural' connections and mathematical abstractions that these connections generate (e.g., representing combinations of pixels in an image) \cite{bansal_challenge_nodate}. This has earned deep neural networks a reputation of being ``black boxes'', an apparatus whose inner-workings remain opaque to the outside observer. The lack of transparency and intelligibility challenges their use for high-stakes decision-making. Notably, black-box models (1) lack quality assurance, (2) fail to elicit trust, and (3) restrict clinician-patient dialogue.

\subsection{Black-boxes can lack quality assurance}

\subsubsection{Errors}

Without a sufficient understanding of \textit{how} a model works and generates predictions, it becomes very difficult to detect errors in a model's performance. Such errors could be introduced incidentally, for example through mislabelled/incomplete training data, or they could also be introduced deliberately, for example through adversarial/Trojan attacks \cite{kelly_key_2019}.
Even when errors are detected, as would happen in the case of catastrophic failure, it can be impossible to debug the cause \cite{carabantes_black-box_2020}. When simple errors go undetected by human supervisors, purportedly powerful models become less useful in practice  \cite{rudin_stop_2019}. The issue of quality assurance has received much attention outside of medicine, notably by the U.S. Department of Defense \cite{gunning_darpas_2019}, though the concerns raised here apply to medicine too.

\subsubsection{Biases}

Black-box models also make it difficult to identify and correct model biases \cite{caruana_intelligible_2015}. This includes those that result in disproportionately more errors among patients who belong to underrepresented or marginalized groups \cite{obermeyer_dissecting_2019} (c.f., ``uncertainty bias'' \cite{goodman_european_2017}). Such biases could perpetuate inequities within the existing healthcare system, or worsen outcomes for more vulnerable patients. Some biases arise from incomplete predictor data. For example, one model trained to triage patients with pulmonary infections learned to assign low risk to asthmatics because the model was not conditioned on the intensity of treatment that those very ill patients received (treatments that were essential to their good outcomes) \cite{caruana_intelligible_2015}.
Other biases arise from incomplete sample data that translates poorly to real-world settings. For example, a model trained on a cohort of hospitalized patients may not generalize to a community clinic where the underlying patient distribution differs (also called distributional shift or covariate shift).
There is also concern that some biases could lead to a disastrous ``self-fulfilling prophecy'' scenario. For example, if a model mistakenly deems medical treatment futile, and thus a patient receives no treatment, then a model could \textit{cause} the poor outcome it predicted \cite{challen_artificial_2019}. This in turn reinforces the model's evaluation of futility.

That is, a vicious and harmful feedback loop may result from some AI-based decisions \cite{oneil_weapons_2016}. Of note, Obermeyer et al. identified an important example of a harmful AI bias in medicine which involved an algorithm falsely predicting that Black patients required fewer medical resources than White patients \cite{obermeyer_dissecting_2019}. Although its designers attempted to eliminate racial categorizing, a racial bias nevertheless crept into the algorithm due to historical healthcare costs being used as a proxy for medical need. This proxy generated unfair bias and thus potential harm, since Black patients historically received fewer health resources owing to systemic discrimination -- a problem that the algorithm would tend to compound.     

The problem of bias is hard to avoid. On one hand, most training data are imperfect because learning is done with the data one has, not the sufficiently representative, rich, and accurately labelled data one wants \cite{caruana_intelligible_2015}. On the other hand, even a theoretically fair model can be biased in practice due to how it interacts with the larger healthcare system \cite{decamp_latent_2020}.  Yet, a transparent model at least offers a chance to detect biases when they occur, giving the human-in-the-loop an opportunity to correct them \cite{caruana_intelligible_2015} and prevent the potentially large scale adverse outcomes that black-box AI tools can create.

\subsection{Black-boxes can fail to elicit trust}

\subsubsection{Owing to lack of quality assurance}

The medical profession is built on trust. For medical AI to be successful, it must be trusted by governments, health professionals, and the public. However, black-boxes may be assumed to be accurate and reliable even when they are not or when they are broadly reliable but misfire on occasion. For example, a black-box AI tool may perform well in lab settings but fail, subtly or overtly, in clinical settings. In fact, a recent review highlighted the present deficiency of sound studies supporting the reliability of medical AI applications \cite{topol_high-performance_2019}. A lack of quality assurance could cause models to make errors that either harm patients or fail to provide patients with expected benefits, thereby eroding trust in those specific medical AI tools. Tools that are not trusted may be rejected even when they could (e.g., as a result of further training) provide ample medical benefit. Catastrophic early failures in an AI model could feasibly corrode trust in the idea of medical AI itself before it has become firmly established in healthcare. The partial or complete rejection of potentially beneficial AI medicine by regulators and users should cause concern, since it prevents healthcare from achieving its basic moral and social goals of continually improving the health of individuals and populations while avoiding unnecessary harm \cite{harris_value_1985}. At the limit, a serious erosion of public trust in medical AI as it begins to take hold could even damage trust in healthcare systems themselves \cite{quinn_trust_2020}. Therefore, ensuring the quality and safety of AI tools has an ethical significance that includes and also goes beyond protecting individual patients.

\subsubsection{Owing to lack of interpretability}

A lack of interpretability could also erode trust. An interpretable AI model is one that is intelligible to clinicians and patients and can be sufficiently understood by them. Interpretation offers an important form of validation, and thus in its own right confers trust in the model. To use black-box AI, clinicians would be expected to put trust not only in the equation of the model, but also in the entire database used to train it and in the handling (e.g. labelling) of that database by the designers \cite{rudin_why_2019}. Such a degree of trust may be at odds with how clinicians normally operate, especially when the AI model has not received support from extensive clinical testing. Consulted medical experts are expected to justify their recommendations to their patients before they are acted upon. At times, practitioners are also required to justify their recommendations to their peers and regulating bodies (c.f., https://www.ahpra.gov.au/). The need for justification follows from the professional obligations of practitioners to minimize harm and provide benefit to patients \cite{beauchamp_principles_2001}. Being able to defend and justify medical decisions in intelligible ways is thus an ethical requirement of practitioners which may be hindered by black-box tools.
Although it is technically possible to generate post-hoc explanations for a black-box decision, these explanations can be unreliable or unintelligible (as well as being vulnerable to adversarial perturbation) \cite{alvarez-melis_robustness_2018}. Indeed, such explanations must be inadequate, because otherwise only the explanations, and not the underlying model, would be needed in the first place \cite{rudin_stop_2019}.

The process of distilling an intelligible explanation necessarily involves a loss of information about the true nature of the AI evaluation. Yet the problem is that the underlying evaluation in the black-box is evasive and inadequate to the user's requirements. A growing body of research is exploring the ways in which post-hoc explanations can fail and, correlatively, must be improved \cite{rudin_why_2019,rudin_stop_2019,alvarez-melis_robustness_2018}. For example, Miller shows how the social sciences shed light on the need for human-friendly explanations \cite{miller_explanation_2019}. Amongst other desiderata, human-friendly explanations allow clinicians to contrast the AI-based decision with other possible-but-rejected options, and thereby more satisfactorily account for the actual prediction or decision that was made. The bar that a black-box decision-making system must clear to meet the basic goals of medical practice is then to gain the trust of clinicians and patients by offering an intelligible justification which can be placed under scrutiny and stand up to it \cite{department_of_health_and_social_care_code_2019}.  

\subsection{Black-boxes can restrict the clinician-patient relationship}

\subsubsection{Restriction of clinician input}

The quality of the clinician-patient relationship depends crucially on the the quality of communication and dialogue between these parties \cite{roter_enduring_2000}. When consulted experts justify themselves to patients, they not only typically confer trust in the recommendation, but open up the opportunity for a dialogue in which the justification can be interrogated. These dialogues are important because medicine is qualitative and complex, and even experts can disagree \cite{cabitza_unintended_2017}. The quality of medical advice depends on medical reasoning being open to clarification, probing, and even critique. Quality healthcare delivery also depends on a quality dialogue between experts, which can give the caring clinician new insights about their patient or the patient's condition. This in turn could usefully feed back into the initial clinician-patient dialogue about prognosis and treatment options. Black-boxes that lack sufficient intelligibility do not allow for an iterative knowledge-discovery process \cite{rudin_stop_2019} (though work is underway to develop flexible and interpretable AI systems \cite{payrovnaziri_explainable_2020}). This becomes another way in which the quality of healthcare could be affected by black-boxes. %Hearing a justification may yield a new understanding that gives rise to a new question that lead to further justifications, and so on. The use of a opaque model restricts the scope of the clinician-machine partnership.

Further, there is some risk that dependency on black-boxes that do not or cannot adequately expose their `reasoning' could erode the medical skills and knowledge of practitioners. Indeed, as some argue, dependence may also eventually degrade the ethical skills or moral virtues that practitioners gradually develop by making decisions for themselves and being accountable for them \cite{mittelstadt_ethics_2016}. Some clinicians may be tempted by apparently (but not truly) omnipotent algorithms to practice medicine defensively, deferring unnecessarily to AI decision-makers \cite{grote_ethics_2020}. All such restrictions on clinician input could impair their relationships with patients.    

\subsubsection{Restriction of patient input}

Compared to the clinician, patients are in the more vulnerable position in terms of being exposed to suffering harms, failing to receive vital health benefits, and losing aspects of their autonomous choice. When a model only provides a bare decision, and not also a justification, patients are deprived from having a dialogue about the underlying reasoning. In practice, this could undermine patient autonomy, or the ability of patients to make decisions about healthcare in accordance with their own plans and values. When a model cannot explain its recommendation, patients are required to make healthcare decisions without sufficient information \cite{grote_ethics_2020}. In that case, they are deprived of an opportunity to give the informed consent which is a necessary condition of the exercise of autonomy \cite{vayena_machine_2018}. The use of black-box AI may also obscure when, and if, the model's recommendations contain implied value judgements about the patient's best interests \cite{mcdougall_computer_2019}. For example, a model may recommend a chemotherapy option that increases quantity of life over quality of life (or \textit{vice versa}) without first considering the patient's wishes. Model opacity could thus make it more difficult for a patient and clinician to engage in joint decision-making of a kind that is properly respectful of patient autonomy. Without due care being taken with AI and data-driven approaches, the ``patient's body and voice may increasingly be replaced or supplemented by data representations of [their] state of being'' \cite{mittelstadt_ethics_2016}. Yet, the autonomous voice of patients is now recognized as a vital part of the patient-clinician relationship.

%%%%%%%%%%%%%%%%%%%%%%%%%%%%%%%%%%%%%%%%%%%%%%%%%%%%
%%%%%%%%%%%%%%%%%%%%%%%%%%%%%%%%%%%%%%%%%%%%%%%%%%%%
\section{The Ghost of AI Yet to Come}
%%%%%%%%%%%%%%%%%%%%%%%%%%%%%%%%%%%%%%%%%%%%%%%%%%%%
%%%%%%%%%%%%%%%%%%%%%%%%%%%%%%%%%%%%%%%%%%%%%%%%%%%%

Deep learning has already achieved near-human performance in medical image classification, perhaps most notably in the diagnosis of diabetic retinopathy \cite{gulshan_development_2016,sayres_using_2019}. In 2016, the ``godfather of deep learning'' Geoffrey Hinton remarked, ``They should stop training radiologists now'' \cite{mukherjee_i_2017} (making 2020 the first without new radiology graduates had we acted on the advice). Such a pronouncement at the present time is hyperbolic. Yet, transformative advances in radiology are only part of what medical AI \textit{could} offer. AI is poised to have a major impact across the healthcare sector, including:

\begin{enumerate}
    \item \textbf{Clinical decision-making:} AI could leverage vast amounts of data to inform diagnosis, prognosis, or treatment \cite{shah_artificial_2019}. These treatment recommendations could even take individual variability into account, and be tailored to patients based on their clinical profile \cite{ashley_towards_2016,lenze_framework_2020}.
    
    \item \textbf{Clinician workflow:} Advances in automated note taking, information retrieval, and medical billing are perhaps most encouraging because they could allow clinicians to spend more quality time with patients in lieu of performing administrative duties \cite{esteva_guide_2019}. Already, AI might have benefited healthcare accessibility through automated language translation services \cite{rajkomar_machine_2019}.
    
    \item \textbf{Direct-to-consumer services:} While some industries will market AI products directly to hospitals and clinicians, we have already begun to see direct-to-consumer products, and more will likely follow \cite{vayena_machine_2018}. Examples might include apps that automatically monitor digital sensors, or suggest diagnoses based on user input \cite{rajkomar_machine_2019}.
    
    \item \textbf{Robotic health:} Advances in robotic AI could improve patient prosthetics \cite{rajkomar_machine_2019} or enable robot-assisted surgery \cite{esteva_guide_2019}, with a possibility that the AI can continuously improve through a machine learning architecture known as reinforcement learning \cite{kaelbling_reinforcement_1996}.
\end{enumerate}

\begin{figure}[H]
\centering
\includegraphics[width=(\textwidth)]{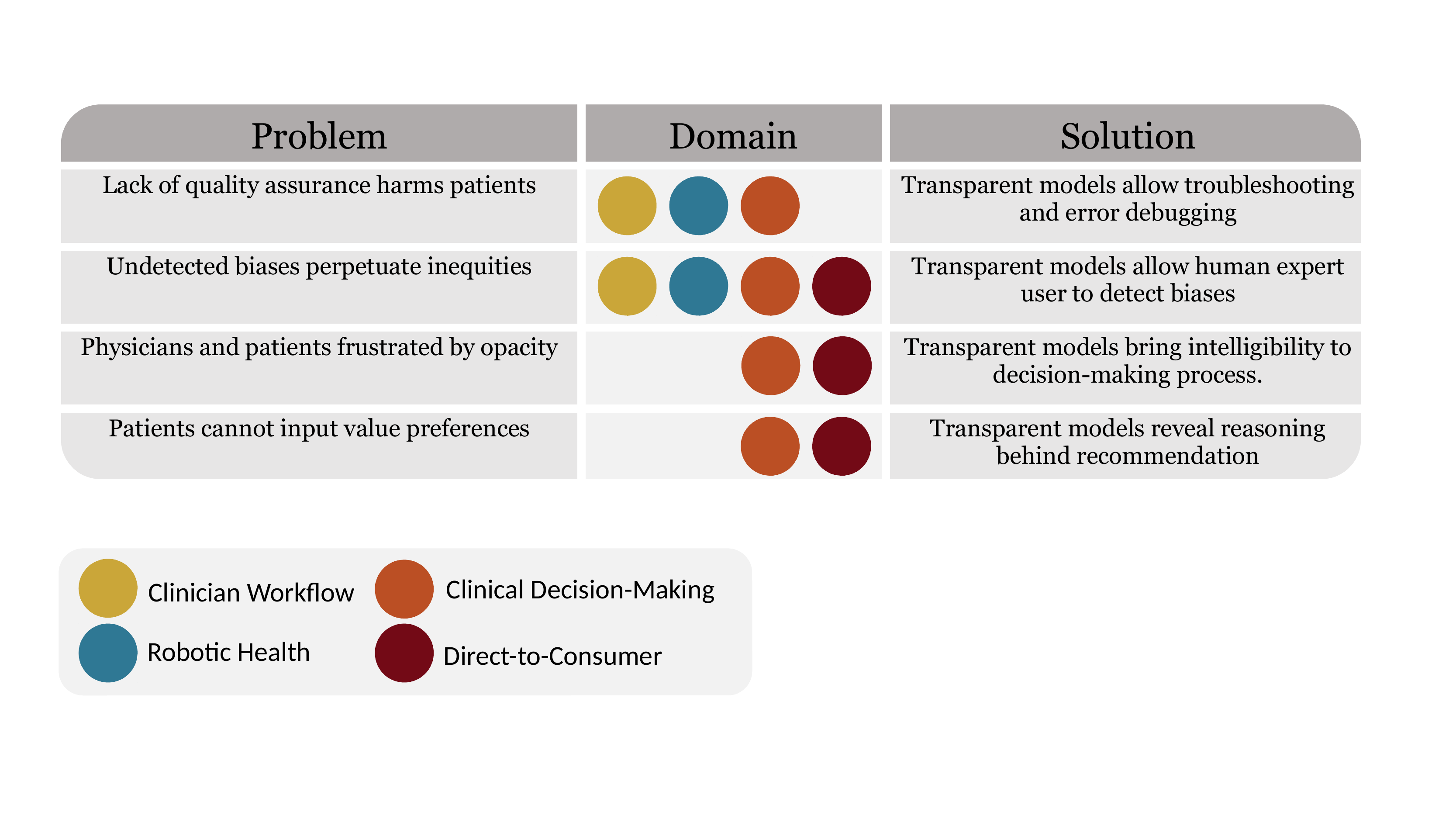}
\caption{AI is poised to have a major impact across the healthcare sector. Black-box models create problems that have the potential to undermine progress in this area. Transparent models offer solutions to these problems.}
\label{fig:consequences}
\end{figure}

Unfortunately, black-box models could feasibly undermine every one of these goals. Figure~\ref{fig:consequences} provides an overview of how each domain -- clinical decision-making, clinician workflow, direct-to-consumer services, and robotic health -- could suffer from the use of models that lack transparency.

\section{A ``clear'' solution: transparency}

Given the list of problems we have raised for black-box tools in medicine, it may seem we are claiming that is simply too risky to introduce them to healthcare systems. However, we are not advocating a miserly aversion to embarking on the AI adventure. On the contrary, our aim in describing some of the darker aspects of black-boxes is to highlight how present AI might be improved in order to ensure that future AI delivers on the benefits that it promises. The key point is that if we are to roll out medical AI successfully, we need transparency. Doctors and patients alike should have reason to trust a model. This requires a change in how we design and validate medical AI to better align with the standards already set for other medical interventions.

% #1) Transparent models
\subsection{Transparent models}

In the case of transparent decision-making, one possible solution is to altogether bypass black-box models of AI—since they depend so heavily on potentially dubious ‘explainability’ (but cf. \cite{miller_explainable_2017})—in favor of intrinsically interpretable AI — which by definition delivers transparent judgments that professionals and patients can easily comprehend \cite{rudin_stop_2019}.
Machine learning models \textit{can} be designed to be intrinsically interpretable. Self-explanatory structures already accompany many models -- including linear regression, decision trees \cite{molnar2020interpretable}, and shallow neural networks \cite{adadi2018peeking} -- where the input, output, and intermediate components have an inherent logical relationship to one another. These intermediate values can be examined to `reverse engineer' the reasoning process that generated the results. In more complex models that use more complex reasoning, the direct interpretation process becomes complicated and quickly expands beyond simple logical comprehensibility (as is the case for many deep neural networks, in which the opacity arises from the multiple layers of non-linearity). However, there are ongoing efforts to make deep models more interpretable, including the thinning of neural connections \cite{le_deep_2020} and the imposing of semantic monotonicity constraints \cite{zhang2018interpretable}.

Adding interpretability may or may not involve sacrificing reliability and accuracy in AI predictions. However, it should involve close partnership with healthcare experts and patients alike who can advise in the design of value-sensitive and value-flexible architectures \cite{mcdougall_computer_2019}. It may also require the design of novel architectures that permit interactive explanation involving the user, say, asking specific questions and receiving answers \cite{miller_explainable_2017,sokol_one_2020}.

In many cases where designing an interpretable model from scratch is impractical, post-hoc explanation methods can examine the behavior of a model after it was constructed. For example, a model can be analyzed by quantifying the contribution of each model component \cite{bau2017network} or feature \cite{beykikhoshk_deeptriage_2020} to the final output decision. The behaviors of units within a deep neural networks can also be explained based on the inputs that trigger them to respond the most \cite{simonyan2014deep}. Alternatively, a deep model can be examined by visualizing knowledge learned by the model \cite{yosinski2015understanding} or studying key data samples used to train them \cite{yin2020dreaming}.
We diverge in opinion from Rudin who argues that there is no place for post-hoc methods in high-stakes decision-making \cite{rudin_stop_2019}. However, we do think that healthcare should, by default, employ intrinsically interpretable models, unless post-hoc explanations can be theoretically justified and empirically validated.

% #2) Transparent validation
\subsection{Transparent validation}
Just because a model is accurate, it does not mean that its use will improve healthcare \cite{keane_eye_2018}. In other words, it is important not to conflate predictive performance with clinical utility \cite{wilkinson_time_2020}. For medical AI to be trusted, it should be clear how it benefits patients. This requires \textit{clinical validation} (as distinct from machine learning validation), which Kim et al. describe as encompassing 3 stages of assessment: (1) diagnostic accuracy in a real-world clinical setting, (2) clinical efficacy, and (3) societal efficacy \cite{kim_design_2019}. Yet, their systematic review found that 94\% of 516 machine learning studies failed to undergo even the first stage of clinical validation \cite{kim_design_2019}, raising concerns that the translational potential of AI may be oversold.

Good clinical validation requires transparency in every aspect of study design and execution, for example by complying with reporting checklists, declaring conflicts of interest, and sharing data and code \cite{montenegromontero_transparency_2019}. AI-specific reporting checklists like CONSORT-AI and SPIRIT-AI may help readers assess the validity and generalizability of medical AI research \cite{consort-ai_and_spirit-ai_steering_group_reporting_2019}, as would more general bias asessment (e.g., PROBAST) and multivariable model reporting (e.g., TRIPOD) checklists \cite{collins_reporting_2019,nagendran_artificial_2020}. Given the prevalence of statistical misconduct due to the pressure to publish, career ambitions, economic motives, and inadequate training \cite{gardenier_misuse_2002}, trial pre-registration may be appropriate \cite{wagenmakers_agenda_2012}. Pre-registration can increase transparency as an ethical obligation, help researchers identify selective reporting, and facilitate peer review \cite{korevaar_facilitating_2017}. The STARD initiative offers advice on where and how to pre-register diagnostic accuracy studies \cite{korevaar_facilitating_2017}.

A transparent model can be evaluated not only in terms of accuracy, but also in terms of its \textit{descriptive accuracy} (i.e., how well the explanation produced by a model captures how the model performs) and \textit{relevancy} (i.e., how well the explanation produced by a model meets the needs of its audience) \cite{murdoch_definitions_2019}.
The clinical validation of interpretable models may require new protocols that can validate the interpretable component of a model specifically \cite{doshi-velez_towards_2017}, for example by testing whether model interpretations support healthcare decision-making as effectively as expert advice.

\section{Conclusion} % delete later....

AI is an increasingly powerful tool that may indeed have the potential to ``revolutionalize'' healthcare one day. However, the mismanagement of AI could have major consequences on public health. Transparency will allow us to adequately monitor whether AI is safe (non-maleficent) and effective (beneficent).
It will also encourage quality assurance, bias detection, and clinician-patient dialogue, all of which will serve to uphold trust in medical AI.
%[IN ORDER TO ACHIEVE THE REAL POTENTIAL OF AI, AND AVOID ANOTHER AI WINTER]

\bibliographystyle{unsrt}
\bibliography{references,vuong,stephan,simon}

\begin{thebibliography}{10}

\bibitem{topol_high-performance_2019}
Eric~J. Topol.
\newblock High-performance medicine: the convergence of human and artificial
  intelligence.
\newblock {\em Nature Medicine}, 25(1):44--56, January 2019.

\bibitem{schwartz_artificial_1987}
W.~B. Schwartz, R.~S. Patil, and P.~Szolovits.
\newblock Artificial intelligence in medicine. {Where} do we stand?
\newblock {\em The New England Journal of Medicine}, 316(11):685--688, March
  1987.

\bibitem{wang_should_2019}
Fei Wang, Rainu Kaushal, and Dhruv Khullar.
\newblock Should {Health} {Care} {Demand} {Interpretable} {Artificial}
  {Intelligence} or {Accept} “{Black} {Box}” {Medicine}?
\newblock {\em Annals of Internal Medicine}, 172(1):59--60, December 2019.

\bibitem{burrell_how_2016}
Jenna Burrell.
\newblock How the machine ‘thinks’: {Understanding} opacity in machine
  learning algorithms.
\newblock {\em Big Data \& Society}, 3(1):2053951715622512, June 2016.

\bibitem{mcculloch_logical_1943}
Warren~S. McCulloch and Walter Pitts.
\newblock A logical calculus of the ideas immanent in nervous activity.
\newblock {\em The bulletin of mathematical biophysics}, 5(4):115--133,
  December 1943.

\bibitem{russell_artificial_2010}
Stuart J. (Stuart~Jonathan) Russell.
\newblock {\em Artificial intelligence : a modern approach}.
\newblock Third edition. Upper Saddle River, N.J. : Prentice Hall, [2010]
  ©2010, 2010.

\bibitem{buchanan_very_2005}
Bruce~G. Buchanan.
\newblock A ({Very}) {Brief} {History} of {Artificial} {Intelligence}.
\newblock {\em AI Magazine}, 26(4):53--53, December 2005.

\bibitem{holzinger_what_2017}
Andreas Holzinger, Chris Biemann, Constantinos~S. Pattichis, and Douglas~B.
  Kell.
\newblock What do we need to build explainable {AI} systems for the medical
  domain?
\newblock {\em arXiv:1712.09923 [cs, stat]}, December 2017.
\newblock arXiv: 1712.09923.

\bibitem{galler_value_1960}
Bernard~A. Galler.
\newblock {THE} {VALUE} {OF} {COMPUTERS} {TO} {MEDICINE}.
\newblock {\em JAMA}, 174(17):2161--2162, December 1960.

\bibitem{haenlein_brief_2019}
Michael Haenlein and Andreas Kaplan.
\newblock A {Brief} {History} of {Artificial} {Intelligence}: {On} the {Past},
  {Present}, and {Future} of {Artificial} {Intelligence}.
\newblock {\em California Management Review}, 61(4):5--14, August 2019.

\bibitem{barnett_computer_1982}
G.~Octo Barnett.
\newblock The {Computer} and {Clinical} {Judgment}.
\newblock {\em New England Journal of Medicine}, 307(8):493--494, August 1982.

\bibitem{hendler_avoiding_2008}
James Hendler.
\newblock Avoiding {Another} {AI} {Winter}.
\newblock {\em IEEE Intelligent Systems}, 23(2):2--4, March 2008.

\bibitem{partridge_scope_1987}
Derek Partridge.
\newblock The scope and limitations of first generation expert systems.
\newblock {\em Future Generation Computer Systems}, 3(1):1--10, February 1987.

\bibitem{jensen_rise_2011}
Lars~Juhl Jensen and Alex Bateman.
\newblock The rise and fall of supervised machine learning techniques.
\newblock {\em Bioinformatics}, 27(24):3331--3332, December 2011.

\bibitem{tu_advantages_1996}
Jack~V. Tu.
\newblock Advantages and disadvantages of using artificial neural networks
  versus logistic regression for predicting medical outcomes.
\newblock {\em Journal of Clinical Epidemiology}, 49(11):1225--1231, November
  1996.

\bibitem{fukushima_neocognitron_1980}
Kunihiko Fukushima.
\newblock Neocognitron: {A} self-organizing neural network model for a
  mechanism of pattern recognition unaffected by shift in position.
\newblock {\em Biological Cybernetics}, 36(4):193--202, April 1980.

\bibitem{waibel_phoneme_1989}
A.~Waibel, T.~Hanazawa, G.~Hinton, K.~Shikano, and K.J. Lang.
\newblock Phoneme recognition using time-delay neural networks.
\newblock {\em IEEE Transactions on Acoustics, Speech, and Signal Processing},
  37(3):328--339, March 1989.

\bibitem{hochreiter_long_1997}
Sepp Hochreiter and Jürgen Schmidhuber.
\newblock Long {Short}-{Term} {Memory}.
\newblock {\em Neural Computation}, 9(8):1735--1780, November 1997.

\bibitem{lecun_deep_2015}
Yann LeCun, Yoshua Bengio, and Geoffrey Hinton.
\newblock Deep learning.
\newblock {\em Nature}, 521(7553):436--444, May 2015.

\bibitem{rudin_why_2019}
Cynthia Rudin and Joanna Radin.
\newblock Why {Are} {We} {Using} {Black} {Box} {Models} in {AI} {When} {We}
  {Don}’t {Need} {To}? {A} {Lesson} {From} {An} {Explainable} {AI}
  {Competition}.
\newblock {\em Harvard Data Science Review}, 1(2), November 2019.

\bibitem{bansal_challenge_nodate}
Daniel S.~Weld Bansal, Gagan.
\newblock The {Challenge} of {Crafting} {Intelligible} {Intelligence}.

\bibitem{kelly_key_2019}
Christopher~J. Kelly, Alan Karthikesalingam, Mustafa Suleyman, Greg Corrado,
  and Dominic King.
\newblock Key challenges for delivering clinical impact with artificial
  intelligence.
\newblock {\em BMC Medicine}, 17(1):195, October 2019.

\bibitem{carabantes_black-box_2020}
Manuel Carabantes.
\newblock Black-{Box} {Artificial} {Intelligence}: {An} {Epistemological} and
  {Critical} {Analysis}.
\newblock {\em AI and Society}, 35(2):309--317, 2020.

\bibitem{rudin_stop_2019}
Cynthia Rudin.
\newblock Stop explaining black box machine learning models for high stakes
  decisions and use interpretable models instead.
\newblock {\em Nature Machine Intelligence}, 1(5):206--215, May 2019.

\bibitem{gunning_darpas_2019}
David Gunning and David Aha.
\newblock {DARPA}’s {Explainable} {Artificial} {Intelligence} ({XAI})
  {Program}.
\newblock {\em AI Magazine}, 40(2):44--58, June 2019.

\bibitem{caruana_intelligible_2015}
Rich Caruana, Yin Lou, Johannes Gehrke, Paul Koch, Marc Sturm, and Noemie
  Elhadad.
\newblock Intelligible {Models} for {HealthCare}: {Predicting} {Pneumonia}
  {Risk} and {Hospital} 30-day {Readmission}.
\newblock In {\em Proceedings of the 21th {ACM} {SIGKDD} {International}
  {Conference} on {Knowledge} {Discovery} and {Data} {Mining}}, {KDD} '15,
  pages 1721--1730, New York, NY, USA, August 2015. Association for Computing
  Machinery.

\bibitem{obermeyer_dissecting_2019}
Ziad Obermeyer, Brian Powers, Christine Vogeli, and Sendhil Mullainathan.
\newblock Dissecting racial bias in an algorithm used to manage the health of
  populations.
\newblock {\em Science}, 366(6464):447--453, October 2019.

\bibitem{goodman_european_2017}
Bryce Goodman and Seth Flaxman.
\newblock European {Union} {Regulations} on {Algorithmic} {Decision}-{Making}
  and a “{Right} to {Explanation}”.
\newblock {\em AI Magazine}, 38(3):50--57, October 2017.

\bibitem{challen_artificial_2019}
Robert Challen, Joshua Denny, Martin Pitt, Luke Gompels, Tom Edwards, and
  Krasimira Tsaneva-Atanasova.
\newblock Artificial intelligence, bias and clinical safety.
\newblock {\em BMJ Quality \& Safety}, 28(3):231--237, March 2019.

\bibitem{oneil_weapons_2016}
Cathy O'Neil.
\newblock {\em Weapons of math destruction: how big data increases inequality
  and threatens democracy}.
\newblock 2016.
\newblock OCLC: 932385614.

\bibitem{decamp_latent_2020}
Matthew DeCamp and Charlotta Lindvall.
\newblock Latent bias and the implementation of artificial intelligence in
  medicine.
\newblock {\em Journal of the American Medical Informatics Association: JAMIA},
  June 2020.

\bibitem{harris_value_1985}
John Harris.
\newblock {\em The {Value} of {Life}}.
\newblock Routledge \& Kegan Paul, 1985.

\bibitem{quinn_trust_2020}
Thomas~P. Quinn, Manisha Senadeera, Stephan Jacobs, Simon Coghlan, and Vuong
  Le.
\newblock Trust and {Medical} {AI}: {The} challenges we face and the expertise
  needed to overcome them.
\newblock {\em arXiv:2008.07734 [cs]}, August 2020.
\newblock arXiv: 2008.07734.

\bibitem{beauchamp_principles_2001}
Tom~L Beauchamp and James~F Childress.
\newblock {\em Principles of biomedical ethics}.
\newblock Oxford University Press, New York, N.Y., 2001.
\newblock OCLC: 758092388.

\bibitem{alvarez-melis_robustness_2018}
David Alvarez-Melis and Tommi~S. Jaakkola.
\newblock On the {Robustness} of {Interpretability} {Methods}.
\newblock June 2018.

\bibitem{miller_explanation_2019}
T.~Miller.
\newblock Explanation in {Artificial} {Intelligence}: {Insights} from the
  {Social} {Sciences}.
\newblock {\em Artif. Intell.}, 2019.

\bibitem{department_of_health_and_social_care_code_2019}
{Department of Health and Social Care}.
\newblock Code of conduct for data-driven health and care technology, 2019.

\bibitem{roter_enduring_2000}
D.~Roter.
\newblock The enduring and evolving nature of the patient-physician
  relationship.
\newblock {\em Patient Education and Counseling}, 39(1):5--15, January 2000.

\bibitem{cabitza_unintended_2017}
Federico Cabitza, Raffaele Rasoini, and Gian~Franco Gensini.
\newblock Unintended {Consequences} of {Machine} {Learning} in {Medicine}.
\newblock {\em JAMA}, 318(6):517--518, August 2017.

\bibitem{payrovnaziri_explainable_2020}
Seyedeh~Neelufar Payrovnaziri, Zhaoyi Chen, Pablo Rengifo-Moreno, Tim Miller,
  Jiang Bian, Jonathan~H. Chen, Xiuwen Liu, and Zhe He.
\newblock Explainable artificial intelligence models using real-world
  electronic health record data: a systematic scoping review.
\newblock {\em Journal of the American Medical Informatics Association},
  27(7):1173--1185, July 2020.

\bibitem{mittelstadt_ethics_2016}
Brent~Daniel Mittelstadt and Luciano Floridi.
\newblock The {Ethics} of {Big} {Data}: {Current} and {Foreseeable} {Issues} in
  {Biomedical} {Contexts}.
\newblock {\em Science and Engineering Ethics}, 22(2):303--341, April 2016.

\bibitem{grote_ethics_2020}
Thomas Grote and Philipp Berens.
\newblock On the ethics of algorithmic decision-making in healthcare.
\newblock {\em Journal of Medical Ethics}, 46(3):205--211, 2020.

\bibitem{vayena_machine_2018}
Effy Vayena, Alessandro Blasimme, and I.~Glenn Cohen.
\newblock Machine learning in medicine: {Addressing} ethical challenges.
\newblock {\em PLoS Medicine}, 15(11), November 2018.

\bibitem{mcdougall_computer_2019}
Rosalind~J. McDougall.
\newblock Computer knows best? {The} need for value-flexibility in medical
  {AI}.
\newblock {\em Journal of Medical Ethics}, 45(3):156--160, March 2019.

\bibitem{gulshan_development_2016}
Varun Gulshan, Lily Peng, Marc Coram, Martin~C. Stumpe, Derek Wu, Arunachalam
  Narayanaswamy, Subhashini Venugopalan, Kasumi Widner, Tom Madams, Jorge
  Cuadros, Ramasamy Kim, Rajiv Raman, Philip~C. Nelson, Jessica~L. Mega, and
  Dale~R. Webster.
\newblock Development and {Validation} of a {Deep} {Learning} {Algorithm} for
  {Detection} of {Diabetic} {Retinopathy} in {Retinal} {Fundus} {Photographs}.
\newblock {\em JAMA}, 316(22):2402--2410, 2016.

\bibitem{sayres_using_2019}
Rory Sayres, Ankur Taly, Ehsan Rahimy, Katy Blumer, David Coz, Naama Hammel,
  Jonathan Krause, Arunachalam Narayanaswamy, Zahra Rastegar, Derek Wu, Shawn
  Xu, Scott Barb, Anthony Joseph, Michael Shumski, Jesse Smith, Arjun~B. Sood,
  Greg~S. Corrado, Lily Peng, and Dale~R. Webster.
\newblock Using a {Deep} {Learning} {Algorithm} and {Integrated} {Gradients}
  {Explanation} to {Assist} {Grading} for {Diabetic} {Retinopathy}.
\newblock {\em Ophthalmology}, 126(4):552--564, 2019.

\bibitem{mukherjee_i_2017}
Siddhartha Mukherjee.
\newblock A.{I}. {Versus} {M}.{D}.
\newblock {\em The New Yorker}, 2017.

\bibitem{shah_artificial_2019}
Pratik Shah, Francis Kendall, Sean Khozin, Ryan Goosen, Jianying Hu, Jason
  Laramie, Michael Ringel, and Nicholas Schork.
\newblock Artificial intelligence and machine learning in clinical development:
  a translational perspective.
\newblock {\em npj Digital Medicine}, 2(1):1--5, July 2019.

\bibitem{ashley_towards_2016}
Euan~A. Ashley.
\newblock Towards precision medicine.
\newblock {\em Nature Reviews Genetics}, 17(9):507--522, September 2016.

\bibitem{lenze_framework_2020}
Eric~J. Lenze, Thomas~L. Rodebaugh, and Ginger~E. Nicol.
\newblock A {Framework} for {Advancing} {Precision} {Medicine} in {Clinical}
  {Trials} for {Mental} {Disorders}.
\newblock {\em JAMA Psychiatry}, 77(7):663--664, July 2020.

\bibitem{esteva_guide_2019}
Andre Esteva, Alexandre Robicquet, Bharath Ramsundar, Volodymyr Kuleshov, Mark
  DePristo, Katherine Chou, Claire Cui, Greg Corrado, Sebastian Thrun, and Jeff
  Dean.
\newblock A guide to deep learning in healthcare.
\newblock {\em Nature Medicine}, 25(1):24--29, January 2019.

\bibitem{rajkomar_machine_2019}
Alvin Rajkomar, Jeffrey Dean, and Isaac Kohane.
\newblock Machine {Learning} in {Medicine}.
\newblock {\em New England Journal of Medicine}, 380(14):1347--1358, April
  2019.

\bibitem{kaelbling_reinforcement_1996}
L.~P. Kaelbling, M.~L. Littman, and A.~W. Moore.
\newblock Reinforcement {Learning}: {A} {Survey}.
\newblock {\em Journal of Artificial Intelligence Research}, 4:237--285, May
  1996.

\bibitem{miller_explainable_2017}
Tim Miller, Piers Howe, and Liz Sonenberg.
\newblock Explainable {AI}: {Beware} of {Inmates} {Running} the {Asylum} {Or}:
  {How} {I} {Learnt} to {Stop} {Worrying} and {Love} the {Social} and
  {Behavioural} {Sciences}.
\newblock {\em arXiv:1712.00547 [cs]}, December 2017.
\newblock arXiv: 1712.00547.

\bibitem{molnar2020interpretable}
Christoph Molnar.
\newblock {\em Interpretable Machine Learning}.
\newblock Lulu. com, 2020.

\bibitem{adadi2018peeking}
Amina Adadi and Mohammed Berrada.
\newblock Peeking inside the black-box: A survey on explainable artificial
  intelligence (xai).
\newblock {\em IEEE Access}, 6:52138--52160, 2018.

\bibitem{le_deep_2020}
Vuong Le, Thomas~P. Quinn, Truyen Tran, and Svetha Venkatesh.
\newblock Deep in the {Bowel}: {Highly} {Interpretable} {Neural}
  {Encoder}-{Decoder} {Networks} {Predict} {Gut} {Metabolites} from {Gut}
  {Microbiome}.
\newblock {\em BMC Genomics}, 21(4):256, July 2020.

\bibitem{zhang2018interpretable}
Quanshi Zhang, Ying Nian~Wu, and Song-Chun Zhu.
\newblock Interpretable convolutional neural networks.
\newblock In {\em Proceedings of the IEEE Conference on Computer Vision and
  Pattern Recognition}, pages 8827--8836, 2018.

\bibitem{sokol_one_2020}
Kacper Sokol and Peter Flach.
\newblock One {Explanation} {Does} {Not} {Fit} {All}.
\newblock {\em KI - Künstliche Intelligenz}, 34(2):235--250, June 2020.

\bibitem{bau2017network}
David Bau, Bolei Zhou, Aditya Khosla, Aude Oliva, and Antonio Torralba.
\newblock Network dissection: Quantifying interpretability of deep visual
  representations.
\newblock In {\em Proceedings of the IEEE conference on computer vision and
  pattern recognition}, pages 6541--6549, 2017.

\bibitem{beykikhoshk_deeptriage_2020}
Adham Beykikhoshk, Thomas~P. Quinn, Samuel~C. Lee, Truyen Tran, and Svetha
  Venkatesh.
\newblock {DeepTRIAGE}: interpretable and individualised biomarker scores using
  attention mechanism for the classification of breast cancer sub-types.
\newblock {\em BMC Medical Genomics}, 13(3):20, February 2020.

\bibitem{simonyan2014deep}
Karen Simonyan, Andrea Vedaldi, and Andrew Zisserman.
\newblock Deep inside convolutional networks: Visualising image classification
  models and saliency maps.
\newblock 2014.

\bibitem{yosinski2015understanding}
Jason Yosinski, Jeff Clune, Anh Nguyen, Thomas Fuchs, and Hod Lipson.
\newblock Understanding neural networks through deep visualization.
\newblock {\em ICML Workshop}, 2015.

\bibitem{yin2020dreaming}
Hongxu Yin, Pavlo Molchanov, Jose~M Alvarez, Zhizhong Li, Arun Mallya, Derek
  Hoiem, Niraj~K Jha, and Jan Kautz.
\newblock Dreaming to distill: Data-free knowledge transfer via deepinversion.
\newblock In {\em Proceedings of the IEEE/CVF Conference on Computer Vision and
  Pattern Recognition}, pages 8715--8724, 2020.

\bibitem{keane_eye_2018}
Pearse~A. Keane and Eric~J. Topol.
\newblock With an eye to {AI} and autonomous diagnosis.
\newblock {\em npj Digital Medicine}, 1(1):1--3, August 2018.

\bibitem{wilkinson_time_2020}
Jack Wilkinson, Kellyn~F. Arnold, Eleanor~J. Murray, Maarten~van Smeden, Kareem
  Carr, Rachel Sippy, Marc~de Kamps, Andrew Beam, Stefan Konigorski, Christoph
  Lippert, Mark~S. Gilthorpe, and Peter W.~G. Tennant.
\newblock Time to reality check the promises of machine learning-powered
  precision medicine.
\newblock {\em The Lancet Digital Health}, 0(0), September 2020.

\bibitem{kim_design_2019}
Dong~Wook Kim, Hye~Young Jang, Kyung~Won Kim, Youngbin Shin, and Seong~Ho Park.
\newblock Design {Characteristics} of {Studies} {Reporting} the {Performance}
  of {Artificial} {Intelligence} {Algorithms} for {Diagnostic} {Analysis} of
  {Medical} {Images}: {Results} from {Recently} {Published} {Papers}.
\newblock {\em Korean Journal of Radiology}, 20(3):405--410, March 2019.

\bibitem{montenegromontero_transparency_2019}
Alejandro Montenegro‐Montero and Alberto~L. García‐Basteiro.
\newblock Transparency and reproducibility: {A} step forward.
\newblock {\em Health Science Reports}, 2(3):e117, 2019.

\bibitem{consort-ai_and_spirit-ai_steering_group_reporting_2019}
{CONSORT-AI and SPIRIT-AI Steering Group}.
\newblock Reporting guidelines for clinical trials evaluating artificial
  intelligence interventions are needed.
\newblock {\em Nature Medicine}, 25(10):1467--1468, 2019.

\bibitem{collins_reporting_2019}
Gary~S. Collins and Karel G.~M. Moons.
\newblock Reporting of artificial intelligence prediction models.
\newblock {\em The Lancet}, 393(10181):1577--1579, April 2019.

\bibitem{nagendran_artificial_2020}
Myura Nagendran, Yang Chen, Christopher~A. Lovejoy, Anthony~C. Gordon, Matthieu
  Komorowski, Hugh Harvey, Eric~J. Topol, John P.~A. Ioannidis, Gary~S.
  Collins, and Mahiben Maruthappu.
\newblock Artificial intelligence versus clinicians: systematic review of
  design, reporting standards, and claims of deep learning studies.
\newblock {\em BMJ}, 368, March 2020.

\bibitem{gardenier_misuse_2002}
John~S. Gardenier and David~B. Resnik.
\newblock The misuse of statistics: concepts, tools, and a research agenda.
\newblock {\em Accountability in Research}, 9(2):65--74, June 2002.

\bibitem{wagenmakers_agenda_2012}
Eric-Jan Wagenmakers, Ruud Wetzels, Denny Borsboom, Han L.~J. van~der Maas, and
  Rogier~A. Kievit.
\newblock An {Agenda} for {Purely} {Confirmatory} {Research}.
\newblock {\em Perspectives on Psychological Science: A Journal of the
  Association for Psychological Science}, 7(6):632--638, November 2012.

\bibitem{korevaar_facilitating_2017}
Daniël~A. Korevaar, Lotty Hooft, Lisa~M. Askie, Virginia Barbour, Hélène
  Faure, Constantine~A. Gatsonis, Kylie~E. Hunter, Herbert~Y. Kressel, Hannah
  Lippman, Matthew D.~F. McInnes, David Moher, Nader Rifai, Jérémie~F. Cohen,
  and Patrick M.~M. Bossuyt.
\newblock Facilitating {Prospective} {Registration} of {Diagnostic} {Accuracy}
  {Studies}: {A} {STARD} {Initiative}.
\newblock {\em Clinical Chemistry}, 63(8):1331--1341, 2017.

\bibitem{murdoch_definitions_2019}
W.~James Murdoch, Chandan Singh, Karl Kumbier, Reza Abbasi-Asl, and Bin Yu.
\newblock Definitions, methods, and applications in interpretable machine
  learning.
\newblock {\em Proceedings of the National Academy of Sciences},
  116(44):22071--22080, October 2019.

\bibitem{doshi-velez_towards_2017}
Finale Doshi-Velez and Been Kim.
\newblock Towards {A} {Rigorous} {Science} of {Interpretable} {Machine}
  {Learning}.
\newblock February 2017.

\end{thebibliography}

\end{document}